\documentclass[lettersize,journal]{IEEEtran}
\usepackage{amsmath,amsfonts}
\usepackage{algorithmic}
\usepackage{algorithm}
\usepackage{array}
\usepackage{bm}
\usepackage[caption=false,font=normalsize,labelfont=sf,textfont=sf]{subfig}
\usepackage{textcomp}
\usepackage{stfloats}
\usepackage{url}
\usepackage{verbatim}
\usepackage{graphicx}
\usepackage{cite}

\usepackage{color}

\usepackage{multirow}  
\usepackage{multicol}  
\usepackage{booktabs}  
\usepackage{makecell}  
\usepackage[table,xcdraw]{xcolor}

\usepackage[colorlinks=true,linkcolor=blue,citecolor=blue,urlcolor=blue,]{hyperref} 
\usepackage[doipre={doi:~}]{uri}

\begin{document}

\title{Aquatic-GS: A Hybrid 3D Representation for Underwater Scenes}




\author{Shaohua Liu, Junzhe Lu, Zuoya Gu, Jiajun Li~\IEEEmembership{Member,~IEEE} and Yue Deng,~\IEEEmembership{Senoir Member,~IEEE}     

\thanks{Shaohua Liu is with the School of Astronautics and the Shenyuan Honors College, Beihang University, Beijing 100191, China (e-mail: liushaohua@buaa.edu.cn).}

\thanks{Junzhe Lu, Zuoya Gu and Jiajun Li are with the School of Astronautics, Beihang University, Beijing 100191, China (e-mail: junzhelu.bh@gmail.com; zuoyagu@buaa.edu.cn; jiajunli@buaa.edu.cn).}

\thanks{Yue Deng is with the School of Artificial Intelligence, Beihang University, Beijing 100191, China (e-mail: ydeng@buaa.edu.cn).}%

\thanks{Corresponding author: Yue Deng}

}

\markboth{Journal of \LaTeX\ Class Files,~Vol.~14, No.~8, August~2021}%
{Liu \MakeLowercase{\textit{et al.}}: Aquatic-GS: A Hybrid 3D Representation for Underwater Scenes}

\IEEEpubid{0000--0000/00\$00.00~\copyright~2021 IEEE}

\maketitle


\begin{abstract}


Representing underwater 3D scenes is a valuable yet complex task, as attenuation and scattering effects during underwater imaging significantly couple the information of the objects and the water. This coupling presents a significant challenge for existing methods in effectively representing both the objects and the water medium simultaneously. To address this challenge, we propose Aquatic-GS, a hybrid 3D representation approach for underwater scenes that effectively represents both the objects and the water medium. Specifically, we construct a Neural Water Field (NWF) to implicitly model the water parameters, while extending the latest 3D Gaussian Splatting (3DGS) to model the objects explicitly. Both components are integrated through a physics-based underwater image formation model to represent complex underwater scenes. Moreover, to construct more precise scene geometry and details, we design a Depth-Guided Optimization (DGO) mechanism that uses a pseudo-depth map as auxiliary guidance. After optimization, Aquatic-GS enables the rendering of novel underwater viewpoints and supports restoring the true appearance of underwater scenes, as if the water medium were absent. Extensive experiments on both simulated and real-world datasets demonstrate that Aquatic-GS surpasses state-of-the-art underwater 3D representation methods, achieving better rendering quality and real-time rendering performance with a 410$\times$ increase in speed. Furthermore, regarding underwater image restoration, Aquatic-GS outperforms representative dewatering methods in color correction, detail recovery, and stability. Our models, code, and datasets can be accessed at \href{https://aquaticgs.github.io}{https://aquaticgs.github.io}.

\end{abstract}

\section{Introduction}


\begin{figure}[!t]
\centering
\includegraphics[width=1.0\columnwidth]{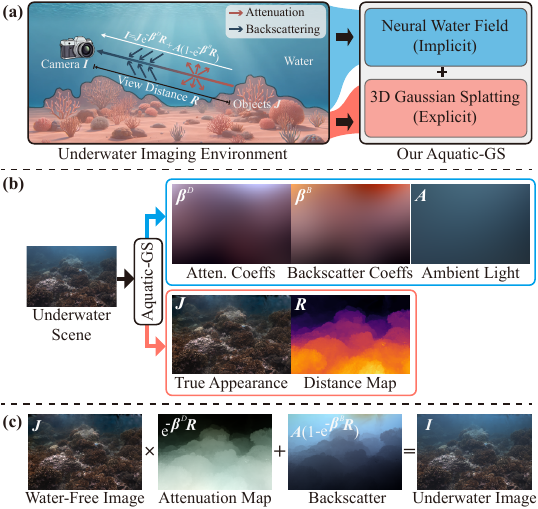}
\caption{(a) Underwater imaging environment and the hybrid representation strategy employed by Aquatic-GS. (b) Scene information learned by Aquatic-GS, including water parameters, the true appearance, and the geometry of the underwater scene. 'Atten.' and 'Coeffs' are abbreviations for Attenuation and Coefficients, respectively. (c) Rendering of an underwater image by Aquatic-GS using a physics-based underwater imaging model.}
\label{fig_imageformation}
\end{figure}

\IEEEPARstart{T}{he} 3D representation of underwater scenes plays a vital role in a wide range of applications, including autonomous underwater vehicles (AUVs)~\cite{wang2023real}, marine ecosystem studies~\cite{yuval2024releasing}, and underwater virtual reality systems~\cite{johnson2017high, llorach2023experience}. An effective underwater 3D representation should encompass both objects (i.e., the water-free scene) and the water medium. The objects provide critical information about the scene's appearance and geometry, while the water medium conveys essential water parameters. However, achieving this goal remains difficult for vision-based 3D representation methods due to the impact of distance-dependent and wavelength-selective attenuation and scattering inherent in the underwater imaging process~\cite{boittiaux2024sucre}. As illustrated in Fig.~\ref{fig_imageformation}(a), attenuation occurs when light reflected from objects is absorbed by the water as it travels toward the camera, with red light being attenuated more than blue and green, leading to a color cast in captured images. Scattering, particularly backscattering, happens when underwater ambient light is scattered toward the camera by particles suspended in the water, leading to a hazy appearance and low contrast in underwater images. Moreover, these two water effects become more pronounced as the distance between the object and the camera increases, leading to a significant coupling of the information from both the objects and the water medium. Such coupling complicates the modeling of both the objects and the water medium, hindering the capture of the true appearance, geometry, and water parameters in underwater scenes~\cite{boittiaux2024sucre}.

\IEEEpubidadjcol  

Recently, advanced 3D scene representation methods, such as Neural Radiance Field (NeRF)~\cite{mildenhall2021nerf} and 3D Gaussian Splatting (3DGS)~\cite{kerbl20233d}, have received significant attention. NeRF employs an implicit modeling strategy, typically leveraging neural networks to efficiently encode complex scenes, and utilizes volume rendering to produce photo-realistic images. In contrast, 3DGS explicitly models the scene with a set of learnable 3D Gaussian primitives and employs a tile-based rasterization pipeline, resulting in superior rendering quality, reduced training cost, and faster rendering speed compared to NeRF. However, in the context of underwater scene representation, both NeRF and 3DGS struggle to effectively represent both the objects and the water medium simultaneously. For instance, some NeRF-based studies~\cite{zhang2023beyond, levy2023seathru, tang2024neural} extend their volumetric rendering framework to accommodate the water medium; however, their implicit modeling strategies for objects often lead to blurred details, noisy geometry, and significant rendering costs. Conversely, while the explicit modeling strategy employed by 3DGS adeptly represents opaque objects, it struggles with the abundant translucent water medium~\cite{fei20243d}, leading to inevitable artifacts and inaccurate geometry when applied to underwater scenes. 
Thus, in underwater scene representation, how to effectively represent both objects and the water medium simultaneously remains a significant challenge.

To address this challenge, we propose a hybrid 3D representation approach called Aquatic-GS, which combines the advantages of explicit representation and implicit modeling to effectively capture the characteristics of both the objects and the water medium in underwater scenes, as shown in Fig.~\ref{fig_imageformation}.
Specifically, considering the spatial non-uniformity of water properties in real environments~\cite{bekerman2020unveiling, akkaynak2018revised, nakath2021situ}, we designed a Neural Water Field (NWF) to model the distributions of the water parameters implicitly. Simultaneously, motivated by the efficiency of 3DGS in representing objects, we utilize 3D Gaussians to explicitly capture the true appearance and geometry of the scene. Both components are integrated through a physics-based underwater image formation (UIF) model to represent the underwater scene. Moreover, to tackle the issues caused by water effects in accurately representing the scene, we introduce a Depth-Guided Optimization (DGO) mechanism. The DGO mechanism employs pseudo-depth maps generated by the latest monocular depth estimators like DepthAnything~\cite{yang2024depth} to guide Aquatic-GS in achieving a more precise representation of the scene’s geometry and distant details. Once optimized, Aquatic-GS learns the true appearance, geometry, and distributions of water parameters in underwater scenes. This not only enables real-time rendering of novel underwater viewpoints but also supports restoring the underwater scene as if the water medium were absent. 

We evaluated Aquatic-GS's performance in underwater novel view synthesis (NVS) and underwater image restoration (UIR) on three real-world underwater datasets and our simulated dataset. For the NVS task, Aquatic-GS outperforms state-of-the-art (SOTA) NeRF-based underwater representation methods, achieving superior rendering quality, reduced training time, and a 410$\times$ increase in rendering speed for real-time performance. For the UIR task, Aquatic-GS also surpasses representative dewatering methods in color correction, detail recovery, and stability.




Our contributions can be summarized as follows:
\begin{itemize}
\item{We propose Aquatic-GS, a novel 3D representation method for underwater scenes that implicitly models water parameters and explicitly models the scene’s appearance and geometry. This hybrid setup effectively represents both the water medium and the objects, ensuring a comprehensive depiction of complex underwater scenes.}
\item{We extend the latest 3DGS to underwater environments and introduce a Depth-Guided Optimization mechanism to tackle the obstacles posed by distance-dependent water effects, achieving better geometry details.}
\item{Extensive experiments on underwater novel view synthesis and image restoration tasks demonstrate the effectiveness of Aquatic-GS for underwater scene representation.}
\end{itemize}

The rest of this paper is structured as follows: Section~\ref{sec:related} covers the related work. In Section~\ref{sec:method}, we provide a detailed description of the proposed Aquatic-GS. Section~\ref{sec:experiment} presents extensive experimental results. Finally, Section~\ref{sec:conclusion} concludes the paper.

\section{Related work}\label{sec:related}

\subsection{NeRF-Based Underwater Representation Methods}

NeRF is a volumetric scene representation method that utilizes a deep neural network to encode the radiance field. Its implicit representation strategy has the advantage of efficiently compressing complex scenes. NeRF typically queries the scene's density and color at any given point through a neural network and employs volumetric rendering to produce remarkably photorealistic images. Furthermore, it has been widely applied to various types of scenes, including unbounded~\cite{barron2022mip, muller2022instant}, in-the-wild~\cite{yang2023cross}, large-scale~\cite{xu2023grid}, and scattering environments~\cite{zhang2023beyond, levy2023seathru, tang2024neural, ramazzina2023scatternerf}.

For the 3D representation of underwater scenes, NeRF-based methods typically extend their volumetric rendering framework to accommodate the water medium. NeuralSea~\cite{zhang2023beyond} incorporating different light-transmitting physics for the water medium and the objects to represent underwater 3D scenes. SeaThru-NeRF~\cite{levy2023seathru} assigns colors and densities for objects and water at each spatial point, utilizing a scattering-aware volumetric rendering framework to simulate attenuation and scattering, ultimately supporting the rendering of new viewpoints for underwater scenes and water-free scenes. Tang et al.~\cite{tang2024neural} treat water as a semi-transparent object with uniform density and color throughout the space, and jointly optimize its parameters with those of the scene objects. However, the implicit modeling of objects in existing methods often results in blurred details, noisy geometry, and considerable computational overhead, including lengthy training times and slow rendering speeds. In contrast, our Aquatic-GS introduces a hybrid modeling strategy. Unlike these methods that consider the water medium from a volumetric rendering perspective, we view the water medium from the perspective of underwater imaging models. We leverage the advantages of implicit modeling to learn the spatial distribution of water parameters, and we extend the advanced 3DGS to explicitly represent the objects, thereby enabling more accurate geometry, efficient training, and real-time rendering.




\subsection{3D Gaussian Splatting}

3DGS~\cite{kerbl20233d} represents a significant breakthrough in 3D scene representation and has attracted considerable attention for its superior high-fidelity rendering quality and real-time rendering capabilities compared to NeRF. Subsequent research has extended 3DGS to various challenging scenarios, including unconstrained photo collections~\cite{dahmani2024swag}, sparse views~\cite{zhu2025fsgs,xiong2023sparsegs, li2024dngaussian, paliwal2024coherentgs}, blurry image collections~\cite{chen2024deblur, lee2024deblurring}, large-scale scenes~\cite{lin2024vastgaussian}, and dynamic scenes~\cite{wu20244d}. Several approaches introduce edge~\cite{gong2024eggs}, frequency~\cite{zhang2024fregs}, and effective rank~\cite{hyung2024effective} regularization to address over-reconstruction issues, while others~\cite{zhang2024pixel, ye2024absgs, fang2024mini} focus on improving over-reconstructed Gaussian identification for finer-grained representation. Methods like SparseGS~\cite{xiong2023sparsegs} and DNGaussian~\cite{li2024dngaussian} use monocular depth maps for depth supervision, mitigating overfitting in sparse-view scenarios. Other methods incorporate optical flow~\cite{paliwal2024coherentgs} and normal priors~\cite{turkulainen2024dn} to further enhance 3DGS's geometric representation. Additionally, Chen et al.\cite{chen2024deblur} and Lee et al.\cite{lee2024deblurring} model the formation of motion and defocus blur, enabling the reconstruction of sharp radiance fields from blurry image collections. Despite these advancements, 3DGS's explicit modeling strategy limits its effectiveness in underwater scene representation and restoration~\cite{li2024watersplatting, zhang2024recgs}. In this work, we extend 3DGS to underwater environments by integrating a neural water field and a physics-based underwater image formation model. Our Aquatic-GS accurately captures the scene's true appearance, geometry, and water parameters, enabling novel underwater viewpoint synthesis and supporting reliable scene restoration.

\subsection{Underwater Image Restoration}

The goal of underwater image restoration is to correct the color cast and low contrast caused by water, obtaining the true appearance of underwater scenes. Existing underwater image restoration methods can be broadly classified into four categories: visual prior-based, data-driven, physics-based, and NeRF-based approaches.

Visual prior-based methods~\cite{ancuti2012enhancing, zhang2022underwater, zhou2024pixel} typically adjust pixel values to enhance underwater images but often overlook the underwater imaging mechanism. Data-driven approaches use supervised learning with synthetic datasets~\cite{badran2023daut, li2019underwater, xie2024uveb}, GANs~\cite{wang2023domain}, or contrastive learning~\cite{liu2022twin} to map underwater images to in-air distributions.
%
%
Physics-based methodss~\cite{drews2016underwater, peng2017underwater, akkaynak2019sea, berman2020underwater, fu2022unsupervised, nakath2021situ, boittiaux2024sucre} consider underwater image formation models to estimate parameters and reverse the degradation process. These methods often incorporate priors, such as underwater dark channel~\cite{drews2016underwater}, and haze-lines~\cite{berman2020underwater}, to facilitate parameter estimation, while some~\cite{nakath2021situ, boittiaux2024sucre} leverage multi-view observations to enhance parameter estimation by constructing 3D structures. For instance, SUCRe~\cite{boittiaux2024sucre} employs the 3D structure to track color changes of a point from different viewpoints, resulting in more accurate parameter estimation. However, most rely on simplified image formation models that assume water parameters are globally uniform within each channel, which may lead to instability in practical applications~\cite{bekerman2020unveiling, li2019underwater}.
NeRF-based methods~\cite{sethuraman2023waternerf, levy2023seathru, zhang2023beyond, tang2024neural}, like SeaThru-NeRF~\cite{levy2023seathru}, typically extend volumetric rendering framework to accommodate the water medium and learn the water-free scenes. However, their reliance on implicit modeling results in blurred details, high training costs, and considerably slow rendering speed.
In contrast, our Aquatic-GS extends the latest 3DGS to learn the water-free scene, while integrating a neural water field to learn the distribution of water parameters. This integration better characterizes complex underwater environments, decouples water effects from the true appearance of underwater scenes, and enables reliable underwater image restoration with lower training costs and real-time rendering performance.

\begin{figure*}[!t]
\centering
\includegraphics[width=7in]{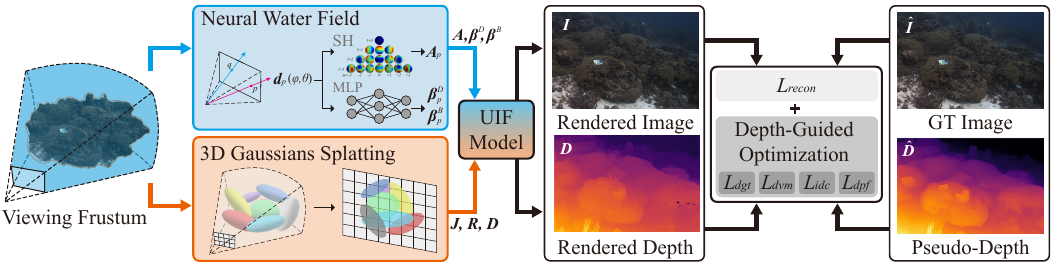}
\caption{Pipeline of Aquatic-GS. For a given viewpoint within a bounded viewing frustum, Aquatic-GS uses the Neural Water Field to obtain the underwater ambient light $\bm{A}$, attenuation coefficients $\bm{\beta}^D$, and backscattering coefficients $\bm{\beta}^B$ for the current viewpoint, while utilizing 3D Gaussian Splatting to render the water-free image $\bm{J}$ along with the corresponding depth $\bm{D}$ and distance maps $\bm{R}$. These outputs are then unified through a physics-based underwater image formation (UIF) model to render the corresponding underwater image $\bm{I}$. During the optimization, along with the reconstruction loss between $\bm{I}$ and $\hat{\bm{I}}$, the Depth-Guided Optimization mechanism, which includes four specifically designed loss functions, leverages a pseudo-depth map $\hat{\bm{D}}$ to guide Aquatic-GS in producing more precise scene representation.}
\label{fig_Framework}
\end{figure*}

\section{methodology}~\label{sec:method}

\subsection{Underwater Image Formation Model}~\label{sec:UIF_Model}   

According to the physics-based underwater image formation (UIF) model proposed by Akkaynak et al.~\cite{akkaynak2018revised}, the relationship between the captured underwater image and the true appearance of the scene can be expressed as follows:
\begin{equation}
\label{uif}
\bm{I}= \bm{J} e^{-\bm{\beta}^D \bm{R}} + \bm{A} (1 - e^{-\bm{\beta}^B \bm{R}}),
\end{equation}
where $\bm{R} \in {\mathbb{R}}^{h \times w \times 1}$ represents the distance between the objects and the camera corresponding to each pixel in the captured image $\bm{I} \in {\mathbb{R}}^{h \times w \times 3}$, with $h$, $w$ and $3$ denoting the height, width, and channels of the image, respectively. $\bm{J} \in {\mathbb{R}}^{h \times w \times 3}$, the water-free image, represents the image of objects that the camera would capture in the absence of the water medium, reflecting the true appearance of the underwater scene. $\bm{A} \in {\mathbb{R}}^{h \times w \times 3}$ denotes the underwater ambient light, which can be considered as the backscatter water color at infinity. $\bm{\beta}^D \in {\mathbb{R}}^{h \times w \times 3}$ and $\bm{\beta}^B \in {\mathbb{R}}^{h \times w \times 3}$ represent the attenuation coefficients and backscatter coefficients of the water medium, respectively, with three channels modeling the wavelength selectivity of these effects.

Typically, the water parameters $\bm{A}$, $\bm{\beta}^D$ and $\bm{\beta}^B$ are assumed to be uniform within each channel in the scene, while being different across channels, simplifying the UIF model to nine unknowns~\cite{akkaynak2019sea, boittiaux2024sucre}. However, some studies suggest that this simplified version may not fully account for factors such as solar directionality~\cite{bekerman2020unveiling}, viewing angles~\cite{akkaynak2018revised}, and local water composition~\cite{nakath2021situ}, among others, which can lead to instabilities and visually unpleasing results~\cite{li2019underwater}. Therefore, in our work, we relax the assumption of uniformity within each channel and attempt to learn the spatially non-uniform distributions of these parameters across each channel.

\subsection{Overview of Aquatic-GS}   



The pipeline of Aquatic-GS is illustrated in Fig.~\ref{fig_Framework}. The distance-dependent water effects cause exponential decay of object information in underwater images as distance increases. Therefore, we set a bounded viewing frustum with a maximum visible distance of $r_{max}$ for each camera to focus on the scene representation within the visible range. For a given viewpoint, we utilize the neural water field to output the water parameters $\bm{A}$, $\bm{\beta}^D$, and $\bm{\beta}^B$. Simultaneously, we employ 3DGS to render the water-free image $\bm{J}$ and corresponding geometric information of the scene, such as depth map $\bm{D}$ and distance map $\bm{R}$. 
As shown in Fig.~\ref{fig_Framework}(c), these components are unified through the physics-based UIF model described in Eq.\eqref{uif}, which computes the water effects and applies them to the water-free image, enabling the rendering of the corresponding underwater image $\bm{I}$. During the optimization, in addition to computing the reconstruction loss between $\bm{I}$ and the ground truth underwater image $\hat{\bm{I}}$, we design a depth-guided optimization mechanism that uses a pseudo-depth map $\hat{\bm{D}}$ to guide Aquatic-GS in generating more accurate underwater scene representation.


In subsequent sections, we will elaborate on the detailed configurations of the neural water field (Section~\ref{sec:NWF}) and the 3DGS (Section~\ref{sec:3DGS}). The depth-guided optimization mechanism will be discussed in Section~\ref{sec:DGO}, and the implementation details will be provided in Section~\ref{sec:DETAILS}.

\subsection{Neural Water Field}~\label{sec:NWF}  

In real scenarios, the water effects are influenced by the solar directionality and viewing angles~\cite{bekerman2020unveiling, akkaynak2018revised}. Thus, we define the distributions of the water parameters within each channel as functions of the viewing direction and design a neural water field to implicitly learn these distributions. As shown in Fig.~\ref{fig_Framework}, for a given camera, the viewing direction corresponding to the $p$-th pixel is denoted as $\bm{d}_p$, and the corresponding underwater ambient light $\bm{A}_p \in {\mathbb{R}}^3$, attenuation coefficient $\bm{\beta}^D_p \in {\mathbb{R}}^3$, and backscatter coefficient $\bm{\beta}^B_p \in {\mathbb{R}}^3$ for this pixel can be queried from the neural water field. For implementation, we use spherical harmonics (SH) to model the underwater ambient light and a shallow multilayer perceptron (MLP) to model the attenuation and backscatter coefficients. Below, we describe these two modeling strategies in detail. 

\subsubsection{SH for Underwater Ambient Light}
Spherical harmonics have been widely applied for modeling low-frequency variations in environmental lightings, such as the sky's illumination distribution~\cite{habel2008efficient}. In this work, we explore their application in underwater scenes. We use learnable SH coefficients to construct a weighted sum of SH basis functions, approximating the complex distribution of underwater ambient light relative to the viewing direction.

Following the standard practice~\cite{fridovich2022plenoxels}, the underwater ambient light $\bm{A}_p$ for a given viewing direction $\bm{d}_p$ can be represented as: 
\begin{equation}
\label{SH}
\bm{A}_p=\sum_{l=0}^{l_{max}} \sum_{m=-l}^{l} \bm{k}_l^m Y_l^m (\bm{d}_p),
\end{equation}
where $\{Y_l^m(\cdot)\}_{l:0 \leq l \leq l_{max}}^{m: -l \leq m \leq l}$ denotes the set of SH basis functions, and $\{\bm{k}_l^m\}_{l:0 \leq l \leq l_{max}}^{m: -l \leq m \leq l}$ represents the corresponding learnable SH coefficients. Each $\bm{k}_l^m \in {\mathbb{R}}^3$ corresponds to the RGB components of the ambient light. The parameter $l$ indicates the degree of the SH basis function. Notably, each SH basis function defines a distribution on the sphere, with higher $l$ corresponding to more complex distribution and greater variation.  The parameter $l_{max}$ represents the highest degree of SH basis functions used in the approximation, with a larger $l_{max}$ incorporating more SH bases, thus improving the capacity to model complex distributions. For computational efficiency, we set $l_{max}=3$.

\subsubsection{MLP for Attenuation and Backscatter Coefficients}
Considering the additional influence of the local composition of the water~\cite{nakath2021situ}, we employ a shallow MLP to model the complex distributions of $\bm{\beta}^D$ and $\bm{\beta}^B$. 
First, for a given viewing direction $\bm{d}_p$, we convert it into a Cartesian unit vector and apply the positional encoding strategy~\cite{mildenhall2021nerf} to obtain a high-dimensional embedding, denoted as $\bm{d}'_p \in {\mathbb{R}}^{27}$. The corresponding $\bm{\beta}^D_p$ and $\bm{\beta}^B_p$ are then computed as follows: 
\begin{equation}
\label{MLP}
(\bm{\beta}^D_p, \bm{\beta}^B_p) = \sigma( F_2( \sigma( F_1(\bm{d}') ) ) ),
\end{equation}
where $F_1(\cdot)$ and $F_2(\cdot)$ represent linear layers with output feature dimensions of 128 and 6, respectively, and $\sigma(\cdot)$ is the Softplus activation function.

By inputting all pixels' viewing directions into the neural water field, we can query the corresponding $\bm{A}$, $\bm{\beta}^D$, and $\bm{\beta}^B$ for the current viewpoint.

\subsection{3D Gaussian Splatting}~\label{sec:3DGS}      
3D Gaussian Splatting employs a collection of 3D Gaussians to represent scenes explicitly. The spatial distribution of each Gaussian is modeled by the equation:
\begin{equation}
\label{3DGS}
G(\bm{x})=e^{- \frac{1}{2} (\bm{x}-\bm{\mu})^T {\bm{\Sigma}}^{-1} (\bm{x}-\bm{\mu})},
\end{equation}
where $\bm{x}$ is a position in world space, $\bm{\mu}$ is the mean position of the Gaussian, and $\bm{\Sigma}$ is the covariance matrix. In practice, $\bm{\Sigma}$ is computed using a scaling matrix $\bm{S}$ and a rotation matrix $\bm{R}$, defined as $\bm{\Sigma}=\bm{R}\bm{S}\bm{S}^T\bm{R}^T$, ensuring positive semi-definiteness. Each Gaussian also includes a set of learnable spherical harmonics coefficients to model view-dependent color and an opacity attribute $o$ that determines its contribution during the blending process.

For efficient rendering, 3DGS utilizes a tile-based differentiable rasterization pipeline. Initially, each 3D Gaussian $G(\bm{x})$ is projected onto the image plane as a 2D Gaussian $G'(\bm{x})$~\cite{zwicker2001ewa}. These are then depth-sorted based on their z-axis coordinates in camera space. The pixel color on the image plane is computed using $\alpha$-blending:
\begin{equation}
\label{alpha_blending}
c(\bm{x}') = \sum_{i=1}^{N_g} c_i \alpha_i \prod_{j=1}^{i-1}(1-\alpha_j), \quad \alpha_i = o_i G'(\bm{x}'), 
\end{equation}
where $\bm{x}'$ is the pixel position in image space, $c_i$ and $\alpha_i$ are the color and opacity of the $i$-th sorted Gaussian at that position, and ${N_g}$ represents the number of Gaussians involved in the blending. The differentiable rasterization process enables the end-to-end optimization of Gaussian parameters, such as $\bm{\mu}$, $\bm{R}$, $\bm{S}$, ${o}$, and SH coefficients, by calculating and backpropagating the gradients of the reconstruction loss (i.e., L1 loss and D-SSIM) between the rendered image and the ground truth.

 
In our work, 3DGS is employed to model the underwater objects, which correspond to the underlying water-free scene. For a given viewpoint, the RGB image rendered by 3DGS corresponds to the water-free image (i.e., the true appearance of the underwater scene) and is denoted as $\bm{J}$.
Alongside $\bm{J}$, we extend the 3DGS to render the accumulated opacity $\bm{o}$, defined as follows: 
\begin{equation}
\label{alpha_opacity}
 \bm{o}(\bm{x}')= \sum_{i=1}^{N_g} \alpha_i \prod_{j=1}^{i-1}(1-\alpha_j),
\end{equation}
Additionally, the corresponding depth map $\bm{D}$ and distance map $\bm{R}$, essential for rendering underwater images, are rendered as follows:
\begin{equation}
\label{alpha_depth}
\bm{D}(\bm{x}')= \begin{cases}
\frac{\sum_{i=1}^{N_g} d_i \alpha_i \prod_{j=1}^{i-1}(1-\alpha_j)}{ \bm{o}(\bm{x}')}, &{\text{if}} \ \bm{o}(\bm{x}') > 0 \\
r_{max}, &{\text{if}} \ \bm{o}(\bm{x}') = 0 
\end{cases}
\end{equation}
\begin{equation}
\label{alpha_distance}
\bm{R}(\bm{x}')= \begin{cases}
\frac{\sum_{i=1}^{N_g} r_i \alpha_i \prod_{j=1}^{i-1}(1-\alpha_j)}{ \bm{o}(\bm{x}')}, &{\text{if}} \ \bm{o}(\bm{x}') > 0 \\
r_{max}, &{\text{if}} \ \bm{o}(\bm{x}') = 0 
\end{cases}
\end{equation}
Here, $d_i$ is the distance from the $i$-th Gaussian to the camera along the z-axis in camera space, while $r_i$ is the Euclidean distance between the $i$-th Gaussian and the camera. We compute $\bm{D}$ and $\bm{R}$ under two conditions: First, if $\bm{o}(\bm{x}') = 0$, this indicates that no Gaussians contribute to the blending at pixel $\bm{x}'$ within the current viewing frustum. In this case, we set both $\bm{D}(\bm{x}')$ and $\bm{R}(\bm{x}')$ to the frustum's radius, $r_{max}$. Second, if $\bm{o}(\bm{x}') > 0$, we normalize the $\alpha$-blended depth and distance using $\bm{o}(\bm{x}')$.

Once we have obtained the water-free image $\bm{J}$, the distance map $\bm{R}$, and the water parameters $\bm{A}$, $\bm{\beta}^D$, and $\bm{\beta}^B$ generated by the NWF, we can utilize the UIF model described in Eq.\eqref{uif} to synthesize the corresponding underwater image $\bm{I}$, enabling the end-to-end optimization of Aquatic-GS.

\subsection{Depth-Guided Optimization Mechanism}~\label{sec:DGO}   

\begin{figure}[!t]
\centering
\includegraphics[width=1.0\columnwidth]{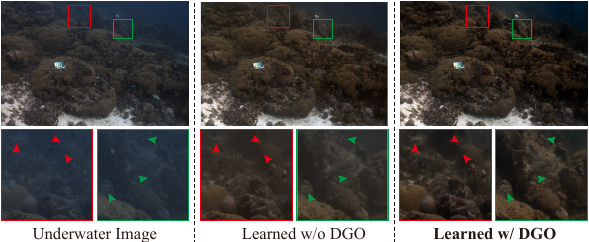}
\caption{Visual comparison of the water-free scene learned with reconstruction loss alone versus those learned using our DGO mechanism. The former displays a haze, especially in distant areas, as highlighted by the red and green boxes. The introduction of our DGO mechanism effectively reduces haze and restores more geometry details in these challenging regions.}
\label{fig_distant}
\end{figure}

Solely using the reconstruction loss does not ensure that Aquatic-GS can effectively decouple the water-free scene from the water effects in underwater images. As shown in Fig.~\ref{fig_distant}, the learned water-free scene using only the reconstruction loss exhibits two main issues: first, the appearance remains hazy, especially in distant areas. Second, in distant, low-contrast areas of the underwater images, the corresponding water-free areas show severe blurring and loss of detail.

These issues primarily stem from the distance-dependent nature of water effects. On one hand, these effects weaken the efficacy of multi-view reconstruction loss in enforcing geometric constraints on Aquatic-GS, especially with increasing distance. Rather than focusing on opaque surfaces, the distribution of 3D Gaussians tends towards a semi-transparent arrangement in space, resulting in a hazy appearance and inaccurate geometry. On the other hand, water effects cause distant areas in underwater images to display significantly weaker texture details and lower contrast than in reality, making it challenging for Aquatic-GS to accurately represent these regions.

To overcome these issues, we have developed a Depth-Guided Optimization (DGO) mechanism, which leverages the pseudo-depth map estimated by DepthAnything~\cite{yang2024depth} to guide Aquatic-GS in accurately representing scene geometry and distant details. This mechanism addresses the issues from four specific aspects: transmittance regularization, minimization of depth variance, coarse-grained depth supervision, and patch-wise frequency domain supervision. Each aspect is implemented through a specifically designed loss function detailed in the following sections.

\subsubsection{Depth-Guided Transmittance Regularization}~\label{sec:DGT}   
Ideally, high-opacity Gaussians should be concentrated on object surfaces, while semi-transparent Gaussians should be minimized~\cite{fang2024mini}. And, for each pixel, the transmittance of Gaussians involved in $\alpha$-blending tends towards 0 or 1, avoiding intermediate values~\cite{reiser2024binary}. To achieve this, we apply transmittance regularization to suppress semi-transparent Gaussians.

First, inspired by~\cite{rebain2022lolnerf}, we introduce a metric $t(\cdot)$ to assess the presence of semi-transparent Gaussians during the $\alpha$-blending process for each pixel as follows:
\begin{equation}
\label{dgo_t}
t(\bm{x}')= \frac{1}{{N_g}} \sum_{i=1}^{N_g} \{ - log ( e^{-10| T_i|} + e^{-10|1 - T_i|} )   \}, 
\end{equation} 
where $\bm{x}'$ represents the coordinates of any given pixel, and $T_i=\prod_{j=1}^{i-1} (1-\alpha_j)$ is the transmittance of $i$-th Gaussian in the pixel' $\alpha$-blending process. A smaller value of $t(\bm{x}')$ indicates that the transmittance values of all the Gaussians contributing to the pixel are closer to either 0 or 1, suggesting fewer semi-transparent Gaussians. We further define a depth-guided transmittance regularization term $L_{dgt}$, calculated as:
\begin{equation}
\label{dgo_dgt}
L_{dgt} = \frac{1}{{HW}} \cdot \sum_{\bm{x}' \in \bm{\Omega} }  \gamma(\bm{x}') t(\bm{x}'),
\end{equation} 
where the entire image domain $\bm{\Omega}$ is segmented into a near region $\bm{\Omega}_{n} = \{ (x,y) | \hat{\bm{D}}(x,y) > \tau_1 \}$ and a far region $\bm{\Omega}_{f} = \{ (x,y)) | \hat{\bm{D}}(x,y) \leq \tau_1 \}$ based on the pseudo-depth map $\hat{\bm{D}}$. The weight $\gamma(\bm{x}')$ is set as $\gamma_n=1$ for $\bm{\Omega}_{n}$ and $\gamma_f=10$ for $\bm{\Omega}_{f}$, with $\gamma_f$ being larger to enhance the suppression of semi-transparent Gaussians in the distant regions.

\subsubsection{Depth Variance Minimization Regularization}~\label{sec:DGMV} 
For any given pixel, the normalized depth, $\bm{D}(\bm{x}')$, computed using Eq.\eqref{alpha_depth}, represents the weighted average depth of all Gaussians participating in the $\alpha$-blending at that pixel. The weight, ${w_i}$, for the $i$-th Gaussian is defined as: ${w_i}=\frac{\alpha_i \prod_{j=1}^{i-1}(1-\alpha_j)}{\sum_{i=1}^{N_g} \alpha_i \prod_{j=1}^{i-1}(1-\alpha_j)}$. Similarly, the variance of depths, $\bm{V}_D(\bm{x}')$, for all Gaussians involved in the $\alpha$-blending process for that pixel can be computed as:
\begin{equation}
\label{dgo_dgmv_var}
\bm{V}_D(\bm{x}')=\sum_{i=1}^{N_g}  {w_i} ( d_i - \bm{D}(\bm{x}'))^2.
\end{equation}

Typically, a smaller value of $\bm{V}_D(\bm{x}')$ implies that the Gaussians contributing to the current pixel are more concentrated. Consequently, we propose a depth variance minimization regularization term, $L_{dvm}$, to encourage a tighter concentration of Gaussians on object surfaces by minimizing the depth variance for each pixel: 
\begin{align} 
L_{dvm} &= \frac{1}{HW} \cdot \sum_{\bm{x}' \in \bm{\Omega}} \eta(\bm{x}') \bm{V}_D(\bm{x}'),
\label{dgo_dgv} 
\end{align} 
where $\bm{\Omega} = \bm{\Omega}_{n} \cup \bm{\Omega}_{f}$ and the weights $\eta(\bm{x}')$ are defined as $\eta_n$ within $\bm{\Omega}_{n} \setminus \bm{\Omega}_{e}$, $\eta_f$ within $\bm{\Omega}_{f} \setminus \bm{\Omega}_{e}$, and $\eta_e$ within ${\bm{\Omega}}_{e}$. The edge region, $\bm{\Omega}_e$, is determined by the pseudo-depth map $\hat{\bm{D}}$. We assign the weights $\eta_n=1$, $\eta_f=0.1$, and $\eta_e=0.001$ to reflect the varying concentration needs across the image.

This depth-guided weighting strategy is motivated by the observation that pixels corresponding to distant regions involve a wider spatial distribution of 3D Gaussians due to perspective projection. To avoid excessive compression of the depth field across this wider spatial distribution, smaller weights are applied. Moreover, in regions identified as edges in the pseudo-depth map, where Gaussians exhibit varied depths, the minimal weight is assigned to prevent excessive smoothing and retain geometric details.


\subsubsection{Inverse Depth Correlation Loss}~\label{sec:IDC}   
Due to the scale ambiguity inherent in the pseudo-depth maps, which encode normalized relative disparity rather than absolute depth values, conventional loss functions like L1 loss do not work well for depth supervision~\cite{paliwal2024coherentgs}. Recent work by Xiong et al.~\cite{xiong2023sparsegs} demonstrated that normalized cross-correlation can effectively measure the similarity between two maps, regardless of discrepancies in their absolute value ranges. Inspired by this, we propose an inverse depth correlation loss function, $L_{idc}$, to facilitate effective depth supervision. We first convert our depth map to its inverse form $\bm{D}' = 1/(\bm{D}+1)$, and then compute the normalized cross-correlation between $\bm{D}'$ and  $\hat{\bm{D}}$. The loss function $L_{idc}$ is defined as follows:
\begin{equation}
\label{dgo_idc}
L_{idc} =1 - {Cov(\bm{D}', \hat{\bm{D}})}/{\sqrt{Var(\bm{D}')Var(\hat{\bm{D}})}}, 
\end{equation}
where $Cov(\cdot, \cdot)$ and $Var(\cdot)$ denote covariance and variance, respectively. This approach not only addresses the issue of scale ambiguity but also enhances the geometric fidelity by aligning $\bm{D}'$ and  $\hat{\bm{D}}$ more closely, providing coarse-grained geometric supervision.


%
\subsubsection{Depth-Guided Patch Frequency Loss}~\label{sec:DPF}   
Frequency domain analysis, adept at capturing subtle details and texture variations, has been widely applied across tasks like image generation~\cite{jiang2021focal}, super-resolution~\cite{korkmaz2024training}, and image fusion~\cite{liu2024mm}. In this study, we propose a depth-guided patch frequency loss $L_{dpf}$, which utilizes frequency domain insights to enhance texture detail perception in distant, low-contrast areas of underwater images, thereby enabling Aquatic-GS to model these challenging areas more effectively.

We begin by partitioning the rendered underwater image $\bm{I}$, the ground truth underwater image $\hat{\bm{I}}$, and the pseudo-depth map $\hat{\bm{D}}$ into patches of $k \times k$ pixels using the same grid pattern. This results in $K$ patch triplets, denoted as  $\{ \bm{I}_i, \hat{\bm{I}}_i, \hat{\bm{D}}_i \}_{i=1}^{K}$, where $K=\lfloor {\frac{h}{k}} \rfloor \times \lfloor {\frac{w}{k}} \rfloor$. We then apply the Discrete Fourier Transform (DFT) to the patches from $\bm{I}$ and $\hat{\bm{I}}$. For the $i$-th patch from $\bm{I}$, denoted as $\bm{I}_i$, the DFT is computed as follows:
\begin{equation}
\label{dgo_dft}
\bm{F}_i(u,v) = \sum_{x=0}^{k-1} \sum_{y=0}^{k-1} \bm{I}_i(x,y) \cdot e^{-2\pi ( \frac{ux}{k} +  \frac{vy}{k} )},
\end{equation}
where $(x,y)$ are the spatial coordinates, $\bm{I}_i(x,y)$ represents the pixel value, $(u,v)$ are the frequency coordinates, and $\bm{F}_i(u,v)$ is the computed complex frequency value. We calculate the magnitude of $\bm{F}_i$, $|\bm{F}_i|$. A similar process is performed for each patch of $\hat{\bm{I}}$. We also calculate the average depth in each patch based on $\hat{\bm{D}}$, denoted $\hat{D}_i^{avg}$. The $L_{dpf}$ loss is then computed as:
\begin{equation}
\label{dgo_DPF}
L_{dpf} = \frac{1}{K}  \sum_{i=1}^K \psi_i \| |\bm{F}_i| -  |\hat{\bm{F}}_i| \|_1,
\end{equation}
where $\psi_i$ is a depth-dependent weight, and $\| |\bm{F}_i| -  |\hat{\bm{F}}_i|   \|_1$ represents the L1-norm between the magnitudes of $\bm{F}_i$ and $\hat{\bm{F}}_i$. Since $\hat{\bm{D}}$ corresponds to disparity, $\psi_i$ is defined as $\psi_i = 1 - \hat{D}_i^{avg}$, ensuring that patches corresponding to distant areas are assigned greater weight and attention. The final $L_{dpf}$ loss encourages Aquatic-GS to focus on enhancing texture detail representation in these challenging distant areas.

\subsection{Implementation Details}~\label{sec:DETAILS}   
Given $M$ observed images of an underwater scene, we first employ COLMAP\cite{schonberger2016structure} to calibrate the extrinsic and intrinsic camera matrices. This process also yields a sparse point cloud, which serves as the initialization for 3D Gaussians. The positions of the cameras in world space are denoted as $\{ \bm{O}_i \}_{i=1}^M$, and the point cloud is denoted by $\{ \bm{P}_j \}_{j=1}^{N_p}$, where $N_p$ is the number of points. We then determine the maximum visible distance in the scene as $r_{max} = \lambda \cdot \max_{i,j} \|\bm{O}_i - \bm{P}_j \|$, with $\lambda$ being a scale factor empirically set to 2.

Because the surfaces in underwater scenes can often be considered Lambertian reflectors~\cite{murez2015photometric}, we limit the highest order of SHs used in 3DGS to zero to enable view-independent color rendering. The final loss function used in the optimization is expressed as follows:
\begin{equation}
\begin{aligned} 
L_{final} = &L_{recon} + \lambda_{dgt} L_{dgt} + \lambda_{dvm} L_{dvm}\\
&+ \lambda_{idc} L_{idc} + \lambda_{dpf} L_{dpf},
\end{aligned}
\label{total_loss}
\end{equation}
where $L_{recon}=(1-\lambda_{ssim})L_1 + \lambda_{ssim} L_{D\text{-}SSIM}$ denotes the reconstruction loss between the rendered and ground truth underwater images, and $L_{D\text{-}SSIM}$ and $\lambda_{ssim}$ denote the D-SSIM loss and its weight, respectively. Additionally, $\lambda_{dgt}$, $\lambda_{dvm}$, $\lambda_{idc}$, and $\lambda_{dpf}$ are the weights assigned to the respective loss functions within our DGO mechanism. After optimization, for a given viewpoint, Aquatic-GS not only renders the underwater image but also supports the rendering of the corresponding water-free image through the 3DGS branch, achieving underwater image restoration.

\section{Experiments}~\label{sec:experiment}
In this section, we demonstrate the effectiveness of our proposed Aquatic-GS through extensive experiments. First, we describe the experimental configurations, including the datasets, implementation details, and evaluation metrics, in Section ~\ref{sec:config}. Then, in Section~\ref{sec:performance_NVS} and Section~\ref{sec:performance_UIR}, we compare the performance of Aquatic-GS with state-of-the-art methods for underwater novel view synthesis and underwater image restoration, respectively. Finally, in Section~\ref{sec:ablation} we conduct ablation studies to further analyze the effectiveness of the proposed Aquatic-GS.

\subsection{Experimental Configurations}~\label{sec:config}
\subsubsection{Datasets}
We employed four distinct datasets to assess the performance of our Aquatic-GS, comprising three datasets from actual underwater environments and one simulated dataset.

{\bf{SeathruNeRF}~\cite{levy2023seathru}:} This dataset comprises four real underwater scenes: Curacao, Panama, Japanese Gardens, and IUI3, each reflecting diverse aquatic and imaging conditions. The scene image counts are 21, 18, 20, and 29, respectively. Following~\cite{levy2023seathru}, these images were downsampled to an average resolution of 900$\times$1400 pixels.

{\bf{Seathru}~\cite{akkaynak2019sea}:} We employed the horizontally-viewed D3 and D5 scenes from this dataset, where each scene consists of 68 and 43 images, respectively.
Color charts with known patterns are distributed throughout these scenes, serving as ground truth to evaluate Aquatic-GS' effectiveness in underwater image color correction.

{\bf{In-the-Wild Underwater (IWU) Dataset}:}
We selected two in-the-wild scenes, Coral and Car, from raw, unstructured footage available online to assess Aquatic-GS's performance in general real-world scenarios. 
For each scene, we sampled frames from the videos, yielding 38 and 54 images, respectively.

{\bf{Our Simulated Dataset}:} 
We constructed three simulated underwater scenes based on Blender, each showcasing distinct challenges: S1 exhibits detailed textures and varied colors; S2, a large-scale environment, displays significant low contrast at distant regions; and S3 shows pronounced greenish color distortions. Each scene contains 50 images with a resolution of 720$\times$1280. More construction details can be found in the supplementary materials.

For real-world scenes, following~\cite{akkaynak2019sea}, we applied white balancing and used COLMAP to obtain camera intrinsics and poses. For simulated scenes, these parameters were exported from Blender and converted to COLMAP format. As our focus is on static underwater scenes, we manually annotated masks to exclude dynamic elements, such as divers in the D5 scene and fish in the Coral scene, to avoid interference.


\begin{table*}[]
\caption{Quantitative evaluations of the underwater NVS task on real-world and simulated datasets. For each dataset, the best metric is highlighted in \textbf{bold} and the second-best in \underline{underlined}. Efficiency metrics are also included.}
\label{tab_nvs}
\renewcommand{\arraystretch}{1.6}  
\setlength{\tabcolsep}{2.6pt}
\centering

\begin{tabular}{l|ccc|ccc|ccc|ccc|c|c}
\hline
\multicolumn{1}{c|}{\multirow{2}{*}{Methods}} & \multicolumn{3}{c|}{Seathru}                                & \multicolumn{3}{c|}{SeathruNeRF}                            & \multicolumn{3}{c|}{IWU}                                    & \multicolumn{3}{c|}{Simulated}                              & \multirow{2}{*}{\begin{tabular}[c]{@{}c@{}}Speed \\ (FPS)\end{tabular}} & \multirow{2}{*}{\begin{tabular}[c]{@{}c@{}}Avg. \\ Time\end{tabular}} \\ \cline{2-13}
\multicolumn{1}{c|}{}                         & PSNR $\uparrow$    & SSIM $\uparrow$   & LPIPS $\downarrow$ & PSNR $\uparrow$    & SSIM $\uparrow$   & LPIPS $\downarrow$ & PSNR $\uparrow$    & SSIM $\uparrow$   & LPIPS $\downarrow$ & PSNR $\uparrow$    & SSIM $\uparrow$   & LPIPS $\downarrow$ &                                                                         &                                                                             \\ \hline
Mip-NeRF 360~\cite{barron2022mip}             & 21.707             & 0.681             & 0.369              & 27.030             & 0.868             & 0.230              & \underline{25.765} & 0.821             & 0.240              & 26.721             & 0.788             & 0.322              & 0.13                                                                    & 12.8h                                                                       \\
InstantNGP~\cite{muller2022instant}           & 20.847             & 0.695             & 0.376              & 20.817             & 0.729             & 0.381              & 20.582             & 0.779             & 0.240              & 25.490             & 0.855             & 0.213              & 0.44                                                                    & 4.7h                                                                        \\
NeuralSea~\cite{zhang2023beyond}              & -                  & -                 & -                  & 25.412             & 0.709             & 0.461              & -                  & -                 & -                  & 28.340             & 0.751             & 0.371              & 0.05                                                                    & 65.8h                                                                       \\
SeaThru-NeRF~\cite{levy2023seathru}           & 21.267             & 0.657             & 0.409              & \underline{27.493} & \underline{0.874} & 0.226              & 25.119             & 0.804             & 0.268              & 30.475             & 0.853             & 0.222              & 0.13                                                                    & 13.5h                                                                       \\
3DGS~\cite{kerbl20233d}                       & \underline{21.764} & \underline{0.757} & \underline{0.280}  & 24.295             & 0.857             & \underline{0.222}  & 24.527             & \underline{0.872} & \underline{0.146}  & \underline{38.243} & \underline{0.974} & \underline{0.050}  & 134.03                                                                  & 0.4h                                                                        \\
Aquatic-GS                                    & \textbf{23.193}    & \textbf{0.782}    & \textbf{0.235}     & \textbf{28.125}    & \textbf{0.894}    & \textbf{0.176}     & \textbf{26.664}    & \textbf{0.885}    & \textbf{0.136}     & \textbf{40.008}    & \textbf{0.981}    & \textbf{0.031}     & 53.64                                                                   & 1.0h                                                                        \\ \hline
\end{tabular}

\end{table*}



\begin{figure*}[!t]
\centering
\includegraphics[width=2.0\columnwidth]{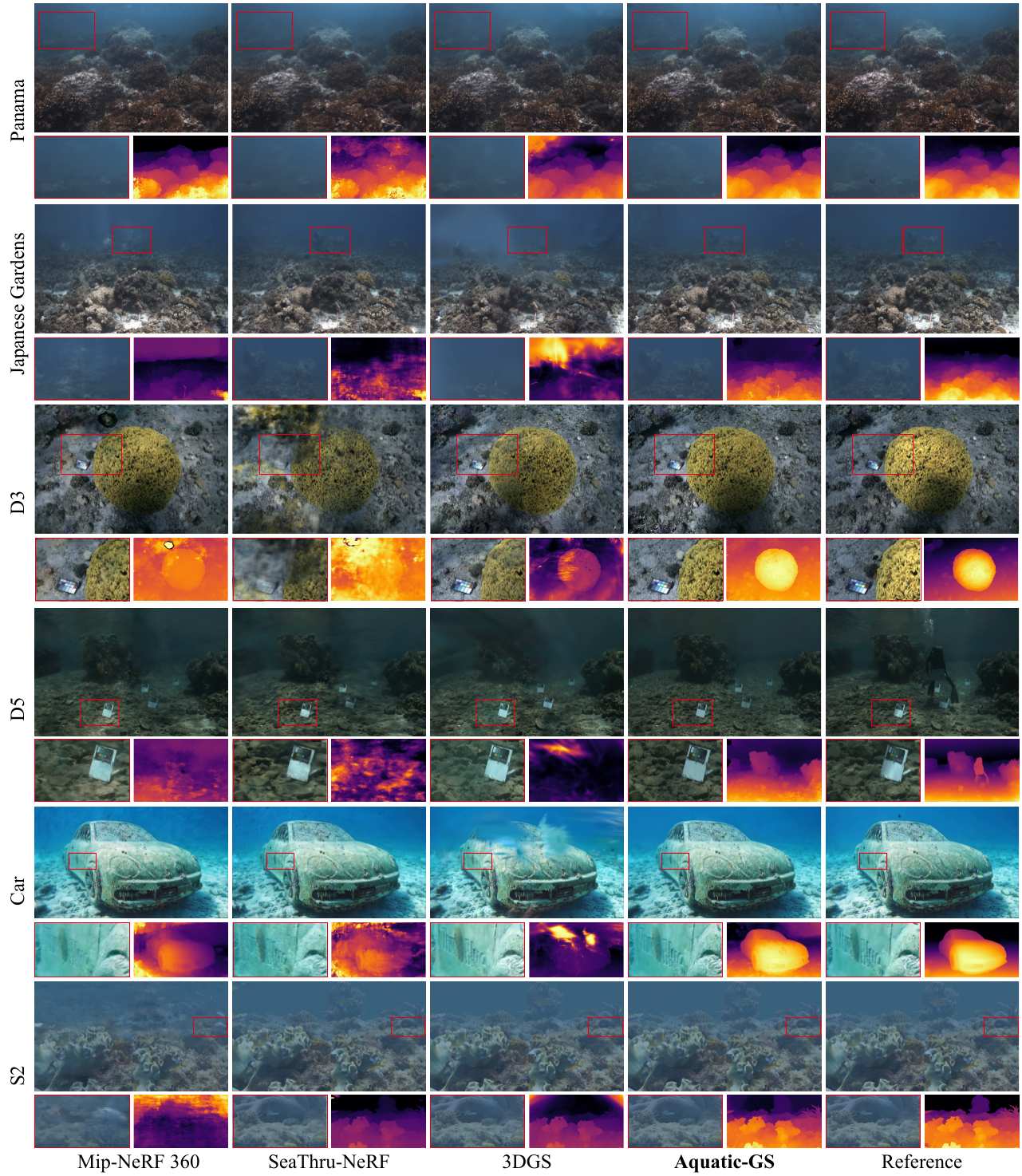}
\caption{Qualitative comparison in the underwater NVS task. Alongside the rendered underwater images, corresponding rendered depth maps are also displayed to assess the geometric representation capabilities of different approaches. Challenging regions are highlighted within red boxes. The pseudo-depth maps serve as benchmarks for evaluating geometric fidelity. Aquatic-GS demonstrates superior rendering quality, effectively reducing blur and artifacts, while reconstructing more details, particularly in distant regions. Regarding geometric representation, our Aquatic-GS generates more reasonable depth maps that are smooth with clear edges and minimal influence from floaters.
}
\label{fig_nvs}
\end{figure*}


\subsubsection{Implementation Details}

Our implementation is based on the official PyTorch version of 3DGS~\cite{kerbl20233d}. We extended its CUDA-based differentiable rasterization pipeline to support the rendering of depth maps, distance maps, and accumulated opacity maps, as well as computations and gradient backpropagation for $L_{dgt}$ and $L_{dvm}$. During training, the weights in Eq.\eqref{total_loss} were $\lambda_{ssim}=0.2$, $\lambda_{dgt}=0.0001$, $\lambda_{dvm}=0.001$, $\lambda_{idc}=0.1$, and $\lambda_{dpf}=0.02$. The patch size $k$ used to compute $L_{dpf}$ is set to 256. For the NWF, we set the learning rates for zero-order SH coefficients at $2.5\times10^{-3}$ and higher-order SH coefficients at $1.25\times10^{-4}$. The MLP within NWF uses an exponential decay learning rate schedule, decreasing from $2\times10^{-3}$ to $2\times10^{-5}$. All other optimization parameters followed the original 3DGS settings. Inspired by ~\cite{zhang2024pixel, ye2024absgs}, we accumulated the per-pixel absolute gradient norms of $\bm{\mu}$ for finer densification. For the NVS task, one in every eight images was selected for testing, with the rest for training. For the UIR task, all images were utilized for training. Each scene underwent 30,000 iterations on a single NVIDIA RTX 3090 GPU.

\subsubsection{Evaluation Metrics}
For the underwater NVS task, we use PSNR, SSIM, and LPIPS metrics to evaluate performance by comparing rendered underwater images of test viewpoints with static ground truth images. We also report the average training time  (Avg. Time) and rendering speed (Frames Per Second, FPS) on an RTX 3090 GPU at a resolution of 720$\times$1280. 

For the UIR task, where the water-free reference image is hard to obtain, color charts provide reliable benchmarks for evaluating color correction. Following recent research~\cite{peng2023u, wang2023domain, bekerman2020unveiling}, we use CIEDE2000 ($\Delta E_{00}$)\cite{gaurav2005ciede2000} and average angular error $\bar{\psi}$ (in degrees)\cite{boittiaux2024sucre} to measure color fidelity. For simulated scenes with clean images, PSNR, SSIM, and LPIPS assess overall restoration quality. Moreover, visual inspection remains essential alongside quantitative metrics.

\subsection{Performance of Underwater Novel View Synthesis}~\label{sec:performance_NVS}

We compared our method with five state-of-the-art approaches, including general scene 3D representation methods Mip-NeRF 360~\cite{barron2022mip}, InstantNGP~\cite{muller2022instant}, and the original 3DGS~\cite{kerbl20233d}, along with two NeRF-based underwater methods: NeuralSea~\cite{zhang2023beyond} and SeaThru-NeRF~\cite{levy2023seathru}. Due to NeuralSea's specific scenario requirements, we evaluated it exclusively on the SeaThru-NeRF dataset and our simulated dataset.

The quantitative results are summarized in Table~\ref{tab_nvs}. Our method consistently demonstrated superior performance across all metrics and datasets, achieving the highest PSNR and SSIM while exhibiting the lowest LPIPS. Notably, the original 3DGS outperformed the NeRF-based approaches overall. On the SeaThru dataset, our method surpassed the second-best 3DGS by 1.429 dB in PSNR. In the SeaThru-NeRF dataset, we improved PSNR by 0.632 dB and SSIM by 0.02 compared to SeaThru-NeRF, while reducing LPIPS by 0.046. In the IWU dataset, Mip-NeRF 360 outperformed 3DGS in PSNR but was still 0.899 dB lower than our method.  On the simulated dataset, our method surpassed 3DGS by 1.765 dB in PSNR. The advantages of our approach over the original 3DGS stem from the distinct modeling of water and objects, as well as the DGO mechanism, demonstrating a more effective scene representation.

Visual comparison results are shown in Fig.~\ref{fig_nvs}. Mip-NeRF 360 struggles to reconstruct distant details, resulting in blurriness. Conversely, SeaThru-NeRF recovers more details in such areas; however, the substantial noise present in the depth maps indicates numerous floaters, suggesting difficulties in learning the true scene geometry. The original 3DGS also struggles to reconstruct details in distant regions significantly affected by water. Its explicit modeling strategy fails to effectively model the water medium, causing numerous floaters to match the training views and resulting in artifacts and incorrect geometry when rendering from test viewpoints. These floaters are closer to the camera than the actual scene objects, shown as bright yellow regions in the depth maps. In contrast, our method achieves superior visual quality with reduced blurriness and artifacts, effectively reconstructing more texture details, particularly in challenging, distant regions. Moreover, it generates a more reasonable depth map, characterized by overall smoothness, sharp edges, and fewer floaters, aligning better with underwater object distribution, demonstrating a more effective scene geometry representation.

In terms of efficiency, we achieve 410$\times$ faster rendering and reduce training time to 7\% of SeaThru-NeRF. Despite the neural water field and DGO mechanism extending training time and slightly reducing speed compared to 3DGS, we maintain real-time rendering at 720$\times$1280 resolution at 53.64 FPS, supporting broad real-time applications.

\begin{table*}[]
\caption{Quantitative evaluations of the UIR task were conducted on three real-world scenes (D3, D5, and Curacao) with color charts and our simulated dataset. Metrics on real-world scenes assess algorithms' color correction capabilities, while those on the simulated dataset evaluate overall underwater image restoration performance. For each scene or dataset, the best metric is highlighted in \textbf{bold} and the second-best in \underline{underlined}.}
\label{tab_uir}
\renewcommand{\arraystretch}{1.6}  
\setlength{\tabcolsep}{7.8pt}
\centering

\begin{tabular}{cl|cc|cc|cc|ccc}
\hline
\multicolumn{2}{c|}{\multirow{2}{*}{Method}}                                                                         & \multicolumn{2}{c|}{D3 Scene}                            & \multicolumn{2}{c|}{D5 Scene}                            & \multicolumn{2}{c|}{Curasao Scene}                       & \multicolumn{3}{c}{Simulated Dataset}                       \\ \cline{3-11} 
\multicolumn{2}{c|}{}                                                                                                & $\Delta E_{00}$ $\downarrow$ & $\bar{\psi}$ $\downarrow$ & $\Delta E_{00}$ $\downarrow$ & $\bar{\psi}$ $\downarrow$ & $\Delta E_{00}$ $\downarrow$ & $\bar{\psi}$ $\downarrow$ & PSNR $\uparrow$    & SSIM $\uparrow$   & LPIPS $\downarrow$ \\ \hline
\multirow{2}{*}{\begin{tabular}[c]{@{}c@{}}Visual\\ Prior-based\end{tabular}} & Fusion~\cite{ancuti2012enhancing}    & 22.94                        & 22.72                     & 24.06                        & 25.60                     & 30.22                        & 26.61                     & 18.152             & 0.873             & 0.097              \\
                                                                              & MLLE~\cite{zhang2022underwater}      & 23.34                        & \underline{21.50}         & 29.97                        & 24.98                     & 23.82                        & 25.20                     & 16.895             & 0.729             & 0.174              \\ \hline
\multirow{4}{*}{Data-driven}                                                  & Waternet~\cite{li2019underwater}     & 21.63                        & 23.39                     & 25.37                        & 24.86                     & 24.86                        & 25.67                     & \underline{20.795} & \underline{0.924} & 0.096              \\
                                                                              & TACL~\cite{liu2022twin}              & 21.59                        & 23.70                     & 25.54                        & 26.62                     & 25.34                        & 26.13                     & 20.219             & 0.909             & 0.123              \\
                                                                              & TUDA~\cite{wang2023domain}           & 24.71                        & 24.52                     & 22.11                        & 25.30                     & 29.00                        & 27.18                     & 19.351             & 0.904             & 0.089              \\
                                                                              & UVEB~\cite{xie2024uveb}              & \underline{19.98}            & 22.18                     & 34.69                        & 26.13                     & \underline{21.03}            & 25.29                     & 19.158             & 0.855             & 0.186              \\ \hline
\multirow{4}{*}{Physics-based}                                                & Seathru~\cite{akkaynak2019sea}       & 23.84                        & 23.16                     & 21.75                        & 23.98                     & 28.43                        & 25.78                     & 19.199             & 0.914             & \underline{0.075}  \\
                                                                              & Hazeline~\cite{berman2020underwater} & 22.49                        & \textbf{20.01}            & 25.11                        & 24.04                     & 30.90                        & 24.63                     & 14.259             & 0.715             & 0.236              \\
                                                                              & USUIR~\cite{fu2022unsupervised}      & 25.08                        & 23.47                     & 23.32                        & 26.29                     & 31.89                        & 27.09                     & 18.902             & 0.883             & 0.124              \\
                                                                              & SUCRe~\cite{boittiaux2024sucre}      & 24.99                        & 23.24                     & \textbf{16.50}               & \underline{21.90}         & 30.58                        & 25.90                     & 20.140             & 0.918             & 0.087              \\ \hline
\multirow{2}{*}{NeRF-based}                                                   & NeuralSea~\cite{zhang2023beyond}     & -                            & -                         & -                            & -                         & 28.17                        & 29.28                     & 11.236             & 0.422             & 0.486              \\
                                                                              & SeaThru-NeRF~\cite{levy2023seathru}  & 20.24                        & 23.57                     & 34.71                        & 26.98                     & 21.36                        & \underline{23.50}         & 15.597             & 0.654             & 0.399              \\ \hline
3DGS-based                                                                    & Aquatic-GS (ours)                    & \textbf{19.95}               & 21.95                     & \underline{20.16}            & \textbf{18.95}            & \textbf{17.73}               & \textbf{21.55}            & \textbf{29.201}    & \textbf{0.970}    & \textbf{0.042}     \\ \hline
\end{tabular}

\end{table*}

\begin{figure}[!t]
\centering
\includegraphics[width=1.0\columnwidth]{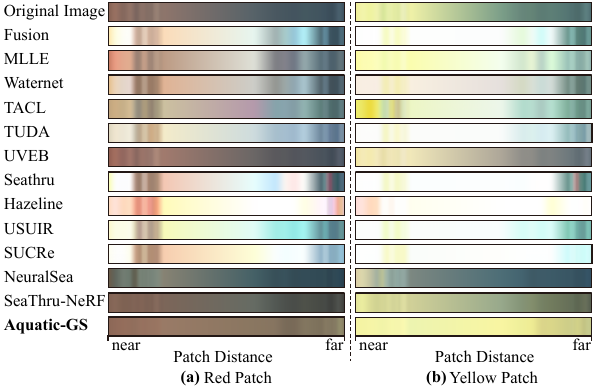}
\caption{Hue values vs. distance. We tracked the hue values of the restored (a) red and (b) yellow color patches at different observation distances within the Curacao scene. The first row illustrates the situation in the original underwater images. The results demonstrate that our Aquatic-GS outperforms other approaches in color correction and maintaining stable hue values.}
\label{fig_hue}
\end{figure}

\subsection{Performance of Underwater Image Restoration}~\label{sec:performance_UIR}

We compared our method with 12 representative underwater image restoration methods, categorized as visual prior-based (Fusion~\cite{ancuti2012enhancing}, MLLE~\cite{zhang2022underwater}), data-driven (Waternet~\cite{li2019underwater}, TACL~\cite{liu2022twin}, TUDA~\cite{wang2023domain}, UVEB~\cite{xie2024uveb}), physics-based (Seathru~\cite{akkaynak2019sea}, Hazeline~\cite{berman2020underwater}, USUIR~\cite{fu2022unsupervised}, SUCRe~\cite{boittiaux2024sucre}), and NeRF-based (NeuralSea~\cite{zhang2023beyond}, SeaThru-NeRF~\cite{levy2023seathru}) approaches. The latter three leverage multi-view information of the underwater scene for 3D reconstruction, while the rest are primarily implemented based on single images. 
  
In terms of color correction accuracy, we analyzed the color differences between restored and expected color patches in three real-world scenes (D3, D5, and Curacao). Table~\ref{tab_uir} shows that our method generally outperforms others in $\Delta E_{00}$ and $\bar{\psi}$ across all scenes. Notably, in the Curacao scene, our approach achieved the lowest $\Delta E_{00}$ and $\bar{\psi}$ values, reducing $\Delta E_{00}$ by 3.3 compared to UVEB and lowering $\bar{\psi}$ by 1.95 degrees relative to SeaThru-NeRF. Moreover, we tracked the hue shifts of the restored red and yellow color patches in the Curacao scene at different observation distances (Fig.~\ref{fig_hue}). Most methods, except UVEB, NeuralSea, and SeaThru-NeRF, oversaturated close-range colors and failed to correct greenish tints at longer distances. While UVEB and SeaThru-NeRF effectively corrected nearby color casts, they struggled with stronger color casts as the patches were farther away. Conversely, our approach consistently delivered effective restoration across various distances, maintaining stable hues and outperforming competitors in color fidelity.

\begin{figure*}[!t]
\centering
\includegraphics[width=2.0\columnwidth]{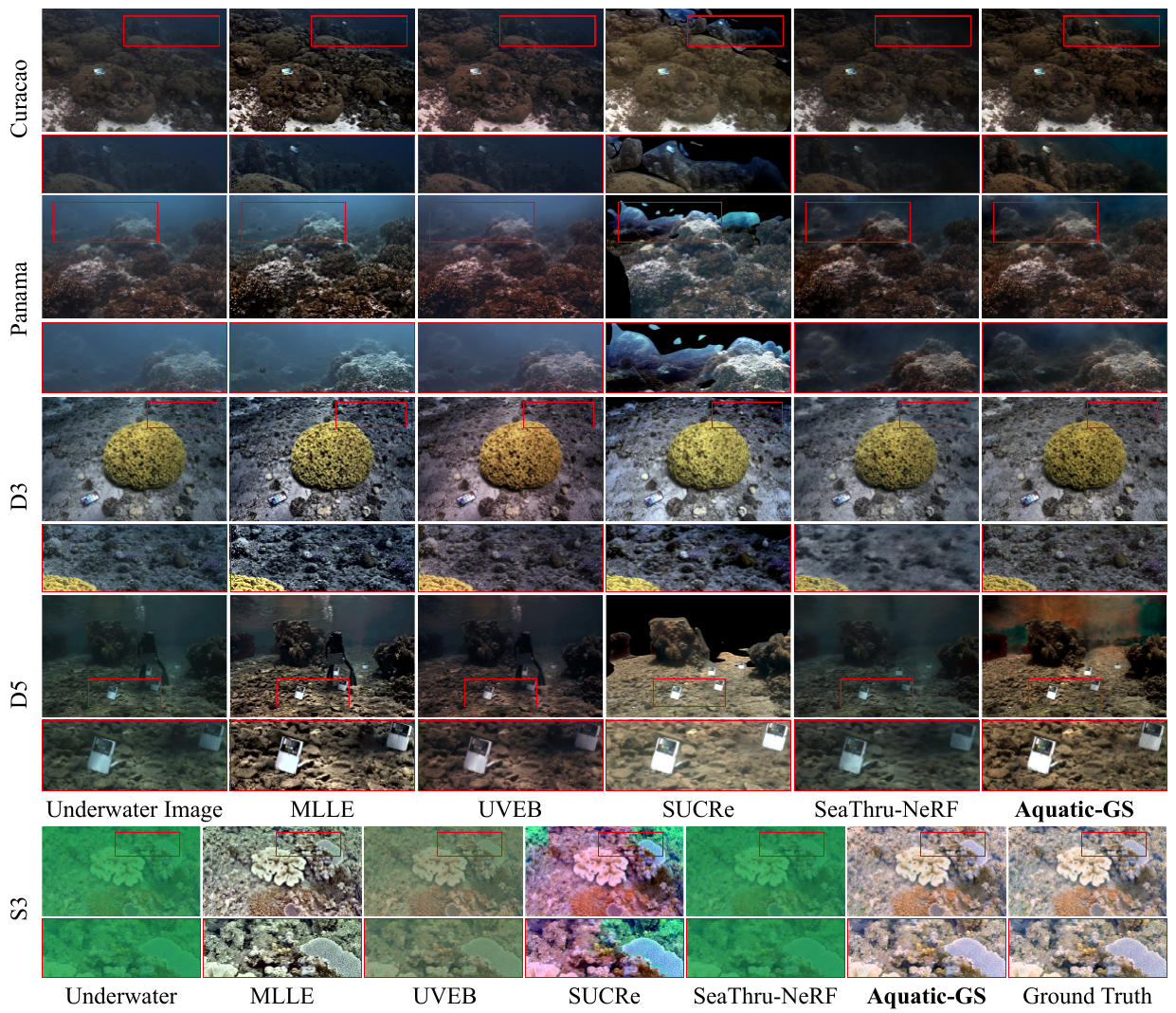}
\caption{Visual comparison in the UIR task. We present visual comparison results among our method and the latest methods from visual prior-based, data-driven, physics-based, and NeRF-based approaches. For the simulated dataset, the corresponding water-free ground truth is displayed. Challenging regions are highlighted with red boxes. Our method effectively restores the true appearance of distant areas, where water effects are pronounced and underwater image contrast is low, achieving more realistic colors and finer texture details. Even in the S3 scene, where the color cast is particularly pronounced, our method yields results that closely match the ground truth.}
\label{fig_uir}
\end{figure*}

For overall performance in underwater image restoration, Table~\ref{tab_uir} shows that our approach outperforms others on the simulated dataset, achieving superior PSNR, SSIM, and LPIPS values—8.406 dB, 0.046, and 0.033 better than the next best methods, respectively—highlighting its effectiveness. Fig.~\ref{fig_uir} provides a visual comparison with state-of-the-art methods. In real-world scenes, MLLE struggles with enhancing distant regions due to a lack of underwater physics consideration. UVEB addresses color deviations partially but leaves a reddish tint and severe backscatter in distant areas, likely due to a domain gap between training and testing images. SUCRe corrects colors and recovers details in distant regions but fails in areas with poorly reconstructed 3D models. SeaThru-NeRF recovers distant appearances more accurately, yet its performance is unstable, with the D5 scene unrecovered, distant regions in the Curacao scene dark, and D3 scene textures blurry. In contrast, our method stably restores scenes with realistic colors and textures, especially in distant regions. In the simulated S3 scene, MLEE corrects color but over-enhances contrast, UVEB and SeaThru-NeRF show limited restoration, and SUCRe removes backscatter but introduces color artifacts. Unlike these, our method effectively mitigates severe water effects, producing results closer to the ground truth. Overall, our method outperforms others in terms of visual fidelity, color correction, detail recovery, and stability.



\subsection{Ablation Studies}~\label{sec:ablation}

To demonstrate the effectiveness of our Aquatic-GS, we conducted a series of ablation studies. These studies systematically evaluated the effectiveness of modeling water parameters with non-uniform distributions, as well as the contribution of each loss function component in the DGO mechanism. All ablation experiments were performed on the SeathruNeRF dataset, with experimental settings kept consistent across all tests.

\begin{table}
\caption{Ablation Study on Modeling Non-uniform Distributions of Water Parameters.}
\label{tab_NWF_setting}
\renewcommand{\arraystretch}{1.6}  
\setlength{\tabcolsep}{2.5pt}
\centering
        
\begin{tabular}{c|cccc}
\hline
Distribution Setting                         & $\Delta E_{00}$ $\downarrow$ & $\Delta E_{00}$ std $\downarrow$ & $\bar{\psi}$ $\downarrow$  & $\bar{\psi}$ std  $\downarrow$\\ \hline
Uniform Distribution     & 19.92           & 4.03                & 21.55          & 2.25             \\
Non-uniform Distribution & \textbf{17.73}  & \textbf{0.64}       & \textbf{21.55} & \textbf{0.97}    \\ \hline
\end{tabular}

\end{table}

\begin{figure}[!t]
\centering
\includegraphics[width=1.0\columnwidth]{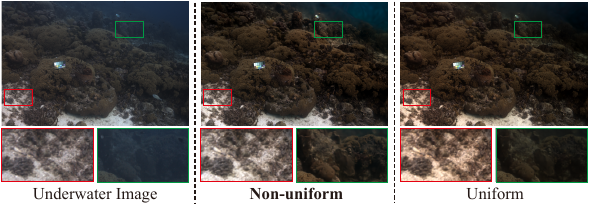}
\caption{Visual comparison of the learned water-free images with different settings of water parameter distribution. The red box highlights a region in the lower left of the view, while the green box emphasizes a region in the upper right. Our non-uniform distribution setting faithfully restores the scene colors, avoiding both over-enhancement and under-restoration.}
\label{fig_HWF}
\end{figure}

\subsubsection{Ablation Study on Modeling Non-uniform Distributions of Water Parameters} 

To validate the need for modeling water parameters as non-uniform distributions, we create a configuration where water parameters are uniformly distributed within each channel. Table~\ref{tab_NWF_setting} quantitatively evaluates color correction performance in the Curacao scene under different water parameter distribution settings. We report the standard deviations of $\Delta E_{00}$ and $\bar{\psi}$ across all restored color charts to assess restoration stability. With uniform water parameter distributions, $\Delta E_{00}$ increased by 2.19, indicating reduced color correction accuracy, while the standard deviations of $\Delta E_{00}$ and $\bar{\psi}$ rose by 3.39 and 1.28, reflecting greater instability.

Fig.~\ref{fig_HWF} shows the learned water-free images under different settings. With the uniform distribution setting, the restored underwater image shows a reddish tint in the lower left view due to over-restoration, while the upper right view appears too dark, indicating insufficient recovery of attenuated light. This suggests that uniformly distributed water parameters are excessive for the lower left and insufficient for the upper right, failing to capture complex spatial variations of water parameters. In contrast, our non-uniform distribution setting provides more reliable restoration, avoiding both over- and under-restoration, demonstrating its effectiveness in complex real-world scenarios.

\begin{table}
\caption{
Ablation Study on the Individual Components of the DGO Mechanism.
}
\label{tab_Ablation_DGO}
\renewcommand{\arraystretch}{1.6}  
\setlength{\tabcolsep}{2pt}
\centering
\begin{tabular}{c|c|c|c|c|ccc}
\hline
Comparative Model & $L_{dgt}$  & $L_{dvm}$  & $L_{idc}$  & $L_{dpf}$  & PSNR $\uparrow$ & SSIM $\uparrow$ & LPIPS $\downarrow$ \\ \hline
Base              &            &            &            &            & 26.695          & 0.879          & 0.205          \\
-                 & \checkmark &            &            &            & 27.092          & 0.880          & 0.205          \\
-                 & \checkmark & \checkmark &            &            & 27.823          & 0.890          & 0.191          \\
-                 & \checkmark & \checkmark & \checkmark &            & 27.906          & 0.893          & 0.190          \\
Aquatic-GS        & \checkmark & \checkmark & \checkmark & \checkmark & \textbf{28.125} & \textbf{0.894} & \textbf{0.176} \\ \hline
\end{tabular}
\end{table}

\subsubsection{Ablation Study on The Individual Components of The DGO Mechanism}

To validate the effectiveness of each loss function component in the proposed depth-guided optimization mechanism, we designate the Aquatic-GS without the DGO mechanism as the Base model. We then incrementally incorporate $L_{dgt}$, $L_{dvm}$, $L_{idc}$, and $L_{dpf}$ to establish various comparison models, as shown in Table~\ref{tab_Ablation_DGO}. This table presents quantitative metrics for different combinations of the DGO mechanism's components in the underwater NVS task, evaluating their performance in representing underwater scenes. The results indicate that each loss function component contributes to performance improvements to varying degrees. Notably, $L_{dgt}$ and $L_{idc}$ significantly enhance the PSNR, while $L_{dvm}$ effectively improves PSNR, SSIM, and LPIPS. The inclusion of $L_{dpf}$ substantially boosts PSNR and reduces LPIPS, resulting in the best performance.

Compared with the Base model, our Aquatic-GS with the DGO mechanism achieves a PSNR improvement of 1.43 dB, an SSIM increase of 0.015, and a further LPIPS reduction of 0.029. As illustrated in Fig.~\ref{fig_distant}, the DGO mechanism effectively facilitates the decoupling of water effects from the water-free scene, leading to a reduction in haze and blur while restoring more geometry details, particularly in challenging distant regions.

\section{Conclusion}~\label{sec:conclusion}

In this paper, we present Aquatic-GS, a novel 3D representation method for underwater scenes. This method employs a hybrid representation strategy that implicitly models the water medium while explicitly representing the objects. Specifically, the neural water field learns the non-uniform distributions of water parameters, while the extended 3DGS accurately captures the true appearance and geometry of underwater scenes. Both components are unified using a physics-based underwater imaging model to generate high-quality underwater images. Our depth-guided optimization mechanism overcomes the negative effects of water, ensuring accurate scene geometry, especially for distant details. Our Aquatic-GS effectively renders novel underwater viewpoints and faithfully restores underwater scenes. Extensive experiments on real-world and simulated datasets demonstrate that Aquatic-GS surpasses SOTA NeRF-based underwater representation methods with higher rendering quality and 410$\times$ faster real-time rendering performance. Furthermore, it outperforms representative dewatering approaches in color correction, detail recovery, and stability.



\bibliography{IEEEabrv,my_article}   
\bibliographystyle{IEEEtran}




\vfill
\end{document}